\documentclass[review]{elsarticle}

\usepackage{lineno,hyperref}
\modulolinenumbers[5]

\usepackage{times}
\usepackage{epsfig}
\usepackage{graphicx}
\usepackage{amsmath}
\usepackage{amssymb}
\usepackage{xspace}
\usepackage{subcaption}
\usepackage{multirow}
\usepackage{color}

\makeatletter
\DeclareRobustCommand\onedot{\futurelet\@let@token\@onedot}
\def\@onedot{\ifx\@let@token.\else.\null\fi\xspace}

\def\eg{\emph{e.g}\onedot} 
\def\ie{\emph{i.e}\onedot}

\def\etal{\emph{et al}\onedot}
\makeatother

\begin{document}

\begin{frontmatter}

\title{MoE-SPNet: A Mixture-of-Experts Scene Parsing Network}


\author[address1]{Huan Fu}
\ead{hufu6371@uni.sydney.edu.au}

\author[address2]{Mingming Gong}
\ead{gongmingnju@gmail.com}

\author[address3]{Chaohui Wang}
\ead{chaohui.wang@u-pem.fr}

\author[address1]{Dacheng Tao}
\ead{dacheng.tao@sydney.edu.au}

\address[address1]{UBTECH Sydney AI Centre, SIT, FEIT, The University of Sydney,
J12 Cleveland St, Darlington NSW 2008, Australia}
\address[address2]{Department of Biomedical Informatics
University of Pittsburgh Cublicle 520c, 5607 Baum Bouevard, Pittsburgh, PA 15206, America}
\address[address3]{Laboratoire d'Informatique Gaspard Monge - CNRS
UMR 8049, Universit$\acute{\text{e}}$ Paris-Est, 77454 Marne-la-Vall$\acute{\text{e}}$e Cedex 2,
France}

\begin{abstract}
Scene parsing is an indispensable component in understanding the semantics within a scene. Traditional methods rely on handcrafted local features and probabilistic graphical models to incorporate local and global cues. Recently, methods based on fully convolutional neural networks have achieved new records on scene parsing. An important strategy common to these methods is the aggregation of hierarchical features yielded by a deep convolutional neural network. However, typical algorithms usually aggregate hierarchical convolutional features via concatenation or linear combination, which cannot sufficiently exploit the diversities of contextual information in multi-scale features and the spatial inhomogeneity of a scene. In this paper, we propose a mixture-of-experts scene parsing network (\emph{MoE-SPNet}) that incorporates a convolutional mixture-of-experts layer to assess the importance of features from different levels and at different spatial locations. In addition, we propose a variant of mixture-of-experts called the adaptive hierarchical feature aggregation (\emph{AHFA}) mechanism which can be incorporated into existing scene parsing networks that use skip-connections to fuse features layer-wisely. In the proposed networks, different levels of features at each spatial location are adaptively re-weighted according to the local structure and surrounding contextual information before aggregation. We demonstrate the effectiveness of the proposed methods on two scene parsing datasets including PASCAL VOC 2012  and SceneParse150 based on two kinds of baseline models FCN-8s and DeepLab-ASPP.
\end{abstract}

\begin{keyword}
Scene Parsing, Mixture-of-Experts, Attention, Convolutional Neural Network
\end{keyword}

\end{frontmatter}


\section{Introduction}

Scene parsing or semantic image segmentation, which predicts a category-level label (such as ``sky", ``dog" or ``person") for each pixel in a scene, is an important component in scene understanding. A perfect parsing can contribute to a variety of applications including unmanned vehicles, environmental reconstruction, and visual SLAM. Many other fundamental computer vision problems can benefit from the parsing of an image, such as medical image analysis, tracking, and object detection \cite{heber2013segmentation, fu2008saliency, leibe2008robust}. However, scene parsing is a very challenging high-level visual perception problem as it aims to simultaneously perform detection, reconstruction, segmentation, and multi-label categorizing \cite{hariharan2014simultaneous,long2015fully}.

Since feature representation is critical to pixel-level labeling problems, classical methods focus on designing handcrafted features for scene parsing \cite{shotton2009textonboost}. Since the handcrafted features alone can only capture local information, probabilistic graphic models such as conditional random fields (CRFs) are often built on these features to incorporate smoothness or contextual relationships between object classes \cite{NIPS2011_4296}. Recently, deep learning approaches such as deep convolutional neural networks (DCNNs) have earned immense success in scene parsing. In particular, fully convolutional networks (FCNs)-based approaches have demonstrated promising performance on several public benchmarks \cite{long2015fully,chen2016deeplab, holder2016road, kampffmeyer2016semantic, deng2015semantic, bulo2017loss}. 

A common strategy adopted in all the CNN-based methods is to aggregate multi-scale/level features from multiple CNN layers \cite{long2015fully} or from a specific layer \cite{chen2016deeplab}, 
which is a key component to obtain high-quality dense predictions because the multi-level features capture different levels of abstractions of a scene. 
The standard way to combine hierarchical features/predictions is to either concatenate multi-level features \cite{hariharan2015hypercolumns, yin2017unsupervised, zhou2016multi, hvrbs2018, passino2010pyramidal, ronneberger2015u, badrinarayanan2017segnet, pohlen2017full} or equivalently aggregate the prediction maps by average pooling \cite{long2015fully}. 
However, the linear feature aggregation methods are not able to evaluate the relative importance of the
semantic and spatial information in each level of features. 
The information at different scales is complementary because
the higher-level convolutional features contain larger-scale contextual information which is beneficial for classification, while the lower-level features have higher spatial resolution which produces finer segmentation masks \cite{ghiasi2016laplacian}. 
The information at different scales is also complementary since they are from different receptive fields.
There is thus a trade-off between the semantic and the spatial information.
In addition, the average pooling ignores the spatial inhomogeneity of a scene, which is improper since different objects may prefer features from different scales/levels. 
For example, textured objects such as ``grass" and ``trees" can be easily distinguished from lower-level features while textureless objects like ``bed" and ``table" require higher-level features to capture the global shape information.

In this paper,  we propose a mixture-of-experts \cite{jordan1994hierarchical} scene parsing network (\emph{MoE-SPNet}) which learns to aggregate multi-level  convolutional features according to the image structures. Specifically, we treat each network branch that contains a specific level/scale of features/predictions as an expert and aggregate them using the weights generated by a trainable convolutional gating network. The gating network also has a convolutional architecture and outputs a weight map for the entire image. 
~The proposed MoE-SPNet is motivated by the following three observations: 1) The lower-level convolutional features contain more precise boundary information but tend to yield more incorrect predictions, while the higher-level features contain more contextual and semantic information but less spatial information.  2) Different levels/scales of features reflect the visual properties of different-sized objects because they are extracted by receptive fields with different sizes. 
Notably, small objects are more likely to be misclassified to their background if using higher-level features because larger receptive fields introduces much noise to these small objects. 
3) The relative importance of different levels of features varies with spatial location; it relies on the local image structure and surrounding contextual information. Obviously, a  linear combination of these features by average pooling cannot capture the homogeneity of a scene and assess the importance of different feature levels. On the contrary, the proposed MoE-SPNet overcomes the limits of linear combination by aggregating different level of features in a nonlinear and adaptive way.

Since MoE-SPNet is only able to adaptively aggregate multi-scale features generated from a single CNN layer, we further propose a variant of MoE called adaptive hierarchical feature aggregation scheme (AHFA) which can be incorporated into the existing parsing networks that aggregate hierarchical features using skip-connections. For example, the original FCN architecture combines features from the last convolutional layer with previous layers by successive upsampling and aggregation. Employing AHFA will enable the parsing networks such as FCN to learn weights at each stage and aggregate the features adaptively as done in MoE-SPNet. In this paper, we focus on exploiting AHFA for the original FCN, leading to a new network architecture denoted as \emph{FCN-AHFA}. 

We demonstrate the effectiveness of our MoE-SPNet and FCN-AHFA on two challenging benchmarks for scene parsing, PASCAL VOC 2012 \cite{everingham2015pascal} and SceneParse150 \cite{zhou2016semantic}, and achieve the state-of-the-art or comparable results. Also, the experimental results show that our MoE-SPNet and FCN-AFHA consistently improve the performance of all the evaluated baseline networks, and thus demonstrate the value of the proposed methods. In addition, the produced weight maps can help us understand the reason that some image structures prefer higher-level convolutional features while others prefer lower-level features.

\section{Related work}

Segmentation is a fundamental problem in scene understanding. While some works focus on low-level segmentation which segments a scene into some regions that share certain characteristics or computed property, such as color, intensity, or texture \cite{shi2000normalized, felzenszwalb2004efficient, liu2010fully, permuter2006study}, high-level segmentation (scene parsing or semantic segmentation), which assigns a category-level label to each pixel of a scene, receives much attention recently.

In the past decade, the successful scene parsing methods rely on handcrafted local features like colour histogram and textons \cite {shotton2009textonboost, bergamasco2012pairwise, xian2015fully, unger2012joint, zemene2016interactive, elguebaly2011nonparametric}, and shallow classifiers such as Boosting \cite{shotton2009textonboost,tu2010auto}, Random Forests \cite{shotton2008semantic, ravi2016semantic}, Support Vector Machines \cite{fulkerson2009class}. Due to the limited discriminative power of local features, a lot of efforts have been put into developing probabilistic graphical models such as CRFs to enforce spatial consistency and incorporate rich contextual information \cite{HeMCR,NIPS2011_4296, chatzis2013conditional, robles2004probabilistic}. Recently, deep learning methods typified by DCNNs have achieved state-of-the-art performance on various computer vision tasks, such as image classification and multi-class object detection. 

Also, the DCNN architectures such as VGG \cite{Simonyan14c} and ResNet \cite{he2016deep} originally developed for image classification have been successfully transferred to scene parsing. Specifically, Long {\it et al.} \cite{long2015fully} proposed the fully convolutional network (FCN) which applied DCNNs to the whole image and directly produced dense predictions from convolutional features, making it possible to get rid of bottom-up segmentation steps \cite{farabet2013learning} and train the parsing network in an end-to-end fashion.

The impressive performance of  FCNs is largely due to the aggregation of multi-level or multi-scale features/predictions. There are mainly two types of aggregation methods: share-nets and skip-nets \cite{CY2016Attention}. The skip-nets, which merge multi-level features/predictions from a single network, are computationally more efficient than the share-nets. Furthermore, they have been refined to enable end-to-end training by normalizing the features from different levels. For example, Hariharan {\it et al.} \cite{hariharan2014simultaneous} concatenated the multi-level features together after certain normalization methods like L2 normalization. However, the concatenation of hierarchical features results in high-dimensional features and is thus time-consuming. The FCN-8s \cite{long2015fully} model aggregated features from the last three convolutional blocks by averagely pooling over layers. Similarly, Chen {\it et al.} \cite{chen2014semantic} 
combined the features which were extracted by applying multi-layer perceptrons on the original image and the pooling layers.
However, linear combination of multi-scale features does not sufficiently exploit the geometric properties, contextual information, and the  spatial-semantic tradeoff. Recently, Ghiasi {\it et al.} \cite{ghiasi2016laplacian} found that directly summing up multi-scale features cannot achieve desirable results, as the learned parameters tended to down-weight the contribution of lower-level features (higher resolution) to suppress the effects of noisy predictions. They proposed the laplacian pyramid refinement approach which computed a boundary mask from higher-level semantic predictions to filter out the noisy predictions in lower-level features. However, we aim to learn the mask weights from multi-level features instead of calculating a boundary mask by manually designed mathematical operations.

Share-nets combine features from shared networks built on multiple rescaled images. For example, Farabet {\it et al.} \cite{farabet2013learning} transformed the raw image through a laplacian pyramid, and each level of which was fed into a CNN. The produced sets of feature maps of all scales were concatenated to form the final representation. Similarly, Lin {\it et al.} \cite{lin2015efficient} resized the original image to three scales and concatenated the multi-scale features. Aside from concatenation, average pooling \cite{dai2015boxsup} and max pooling \cite{Papandreou2014UntanglingLA} were adopted over scales to merge multi-scale features.
However, average or max pooling either treats the multi-scale features equally or losses too much information. Targeting this problem, Chen {\it et al.} \cite{CY2016Attention} proposed the scale attention method which uses the attention model \cite{bahdanau2014neural} over scales to focus on the features from the most relevant scales. Instead of aggregating multi-scale features at one time, Pinheiro {\it et al.} \cite{pinheiro2014} proposed a multi-stage approach which fed multi-scale images successively to a recurrent convolutional neural network. Although the share-nets obtain much better performance, they are computationally more expensive than the single scale networks. Most recently, Chen \etal \cite{chen2016deeplab} developed an atros spatial pyramid pooling strategy (a variant of the share-nets) which extracted multi-scale features in a single network. However, the multi-scale features were still aggregated via an average pooling, and the performance had some gaps against the typical share-nets.

In this paper, we investigate how to adaptively aggregate multi-level or multi-scale features in a single network to further improve their performance and obtain deeper understanding of the special properties of the features from different layers. Specifically, we treat the network branches which obtain multi-level/scale features as  expert networks, and propose MoE-SPNet, which learns some pixel-wise gating weight maps for each experts, to adaptively aggregating these features for a better solution for scene parsing. We also propose the AHFA scheme to further improve existing skip-nets \cite{long2015fully,ghiasi2016laplacian} that use stage-wise aggregation of hierarchical features.  Since most of current parsing networks follow the similar forms as FCN or DeepLab, we can conclude that our technique is widely applicable.

\section{Background}
In this section, we first review the mixture-of-experts (MoE) framework and then review two typical scene parsing networks that employ fully convolutional architectures, i.e., FCN-8s~\cite{long2015fully} and DeepLab-ASPP \cite{chen2016deeplab}.

\subsection{Mixture-of-Experts}
\label{sec:MoE}

\begin{figure}[ht!]

\begin{center}
\begin{subfigure}{1.0\textwidth}
  \begin{center}
  \includegraphics[width=0.6\textwidth]{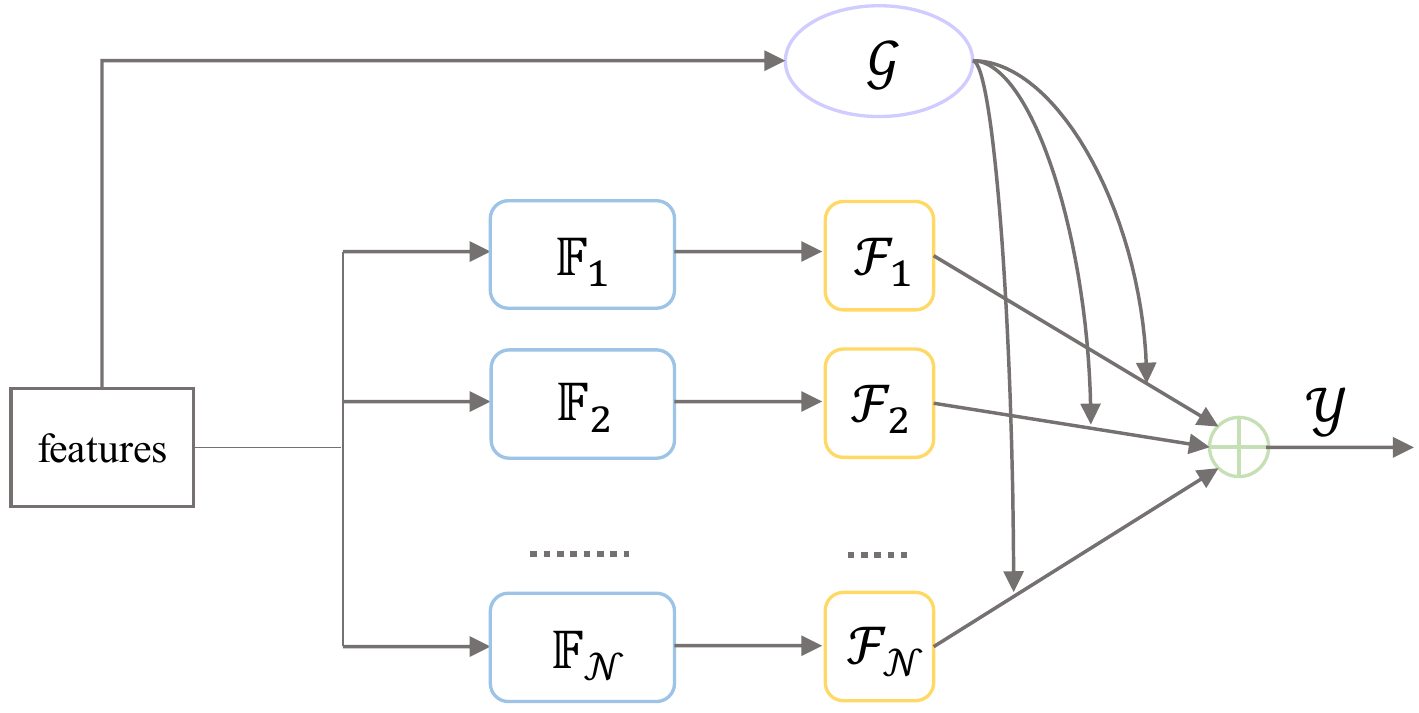}
  \end{center}
\end{subfigure}
\caption{\small{\textbf{Mixture-of-Experts.} An illustration of mixture-of-experts. The same input is fed to different experts, resulting in different solutions for the whole problem space. Typically, the gating network $\mathcal{G}$ also receives the same input as the experts, and the weights are often nomalized by the \emph{softmax} function. \textcolor{black}{Here, $\mathbb{F}_{i}$ and $\mathcal{F}_{i}$ are learned intermediate features and a prediction correspond to expert $i$ respectively.}}}
\end{center}
\label{fig:MoE2}
\end{figure}

Mixture-of-experts \cite{jordan1994hierarchical} is one of the effective machine learning techniques which aims to adaptively aggregating multiple decisions from different experts. As shown in Fig. 1, 
MoE contains two key components: multiple correlated experts and a gating network.  The multiple correlated experts are expected to learn the distribution specialized on a stochastic subspace of the whole problem space, and are thus complementary to each other.  The gating network aims at learning weights for each expert according to their local efficiency. It should be noted that, the weights in the gating network are dynamically determined by the input features. Here, we take mixture-of-expert networks as an example, and introduce the conventional MoE with respect to two different error functions in the learning process. 
\subsubsection{Cooperation Encouraged Error Function}
The error function which encourages cooperation among local experts exhibits the following form: 
\begin{equation}
\begin{split}
E_{coop} = \| y - \sum_{i=1}^{N}{g_{i}o_{i}} \|^{2}
\end{split}
\label{eq:coop}
\end{equation}
where $y$ is the target vector, $N$ is the number of experts, $o_{i}$ is the output of expert $i$, and $g_{i}$ from the gating network ($g_{1} + g_{2} + ... + g_{N} = 1$) represents the contribution of expert $i$ for the final prediction. 

With this error function, the blend of the outputs from each expert is directly compared with the target, meaning that the parameters in each expert are updated according to the overall ensemble error. The strong coupling in the learning process encourages all the experts cooperating nicely, but tends to make each expert generalize to the whole problem space rather than to different subspaces of the  whole problem space. Thus, the learned model via this error function may become inconsistent with the localization of the experts. 

\subsubsection{Competition Encouraged Error Function}
Addressing the shortage in cooperation encouraged error function, Jacobs \etal \cite{jacobs1991adaptive} defined a competition encouraged error function as:
\begin{equation}
\begin{split}
E_{comp} = \sum_{i=1}^{N}{g_{i} \| y - o_{i} \|^{2}}
\end{split}
\label{eq:coop}
\end{equation}

From the definition, this error function actually measures the expected value of differences between the target and each local experts. Thus, each expert directly responds to their own occasions and obtain a complete output over the whole problem space instead of a residual. After the training process, a single expert prefer to generate a solution for a specific training case, and the gating network here plays a role in selecting one or several experts for a given input. In this case, the experts still have some indirect coupling of each other due to the gating network.

\subsection{FCN and DeepLab-ASPP}
FCN-8s \cite{long2015fully} applies a deep convolutional architecture, \eg VGG net \cite{Simonyan14c}, in a fully convolutional fashion to extract hierarchical features with different strides (32x, 16x, and 8x), and combine these features stage by stage from a deeper (coarser) layer to a shallower (finer) layer. Specifically, built on the 16-layer VGG (VGG16) architecture, FCN-8s replaces fully-connected layers with convolutional layers to generate the prediction feature maps with stride 32. 

DeepLab-ASPP reduces the stride of 32x feature maps of FCN to 8 by using dilated convolutions (atrous algorithm) \cite{holschneider1990real}, which introduces zeros to increase the convolution fields for the convolutional kernels. 
Then, the atrous spatial pyramid pooling (ASPP) strategy, which employs multiple parallel filters with different dilation rates on the \texttt{pool5} layer, is adopted to exploit multi-scale features. The generated predictions from the multi-scale features are simply summed together to produce the final prediction.
\section{Approach}
In this section, we first present how to incorporate MoE in a scene parsing network and describe the details of the proposed MoE-SPNet.  Second, we introduce  adaptive hierarchical feature aggregation (AHFA) scheme which is a variant of MoE and show how it can incorporated into the skip-net FCN-8s \cite{long2015fully} to form a new network FCN-AHFA. The AHFA scheme can be incorporated into other skip-nets in a similar way.
\subsection{MoE-SPNet}
\label{sec:MoE-spnet}

\begin{figure*}[ht!]

\begin{center}
\begin{subfigure}{1.0\textwidth}
  \begin{center}
  \includegraphics[scale=0.4]{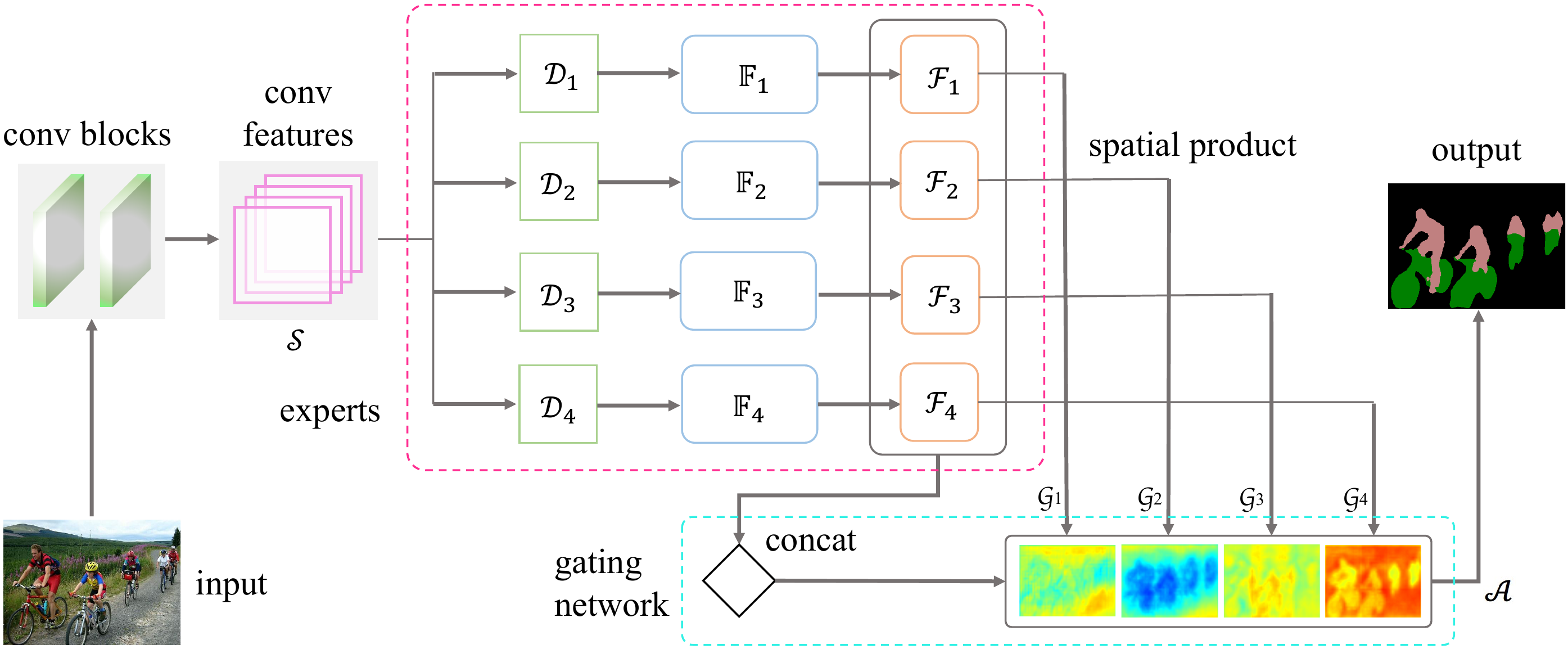}
    \end{center}
\end{subfigure}%
\caption{\small{\textbf{MoE-SPNet.} An illustration of the proposed MoE-SPNet. $\mathcal{D}_{i}$ represents a dilated convolutional layer with a specific dilation rate. We learn 4 experts in this paper with the dilation rates of 6, 12, 18, and 24 respectively. Each expert learns a richer representation ( $\mathbb{F}_{i}$) of the input scene, and produces a solution (denote as $\mathcal{F}_{i}$) for the parsing task. $\mathcal{G}_i$ represents the weight map produced by the gating network for each parsing solution $\mathcal{F}_i$. $\mathcal{A}$ is the final segmentation mask which is the weighted aggregation of all $\mathcal{F}_i$.}}
\label{fig:MoE-spnet}
\end{center}
\end{figure*}

We develop a mixture-of-experts scene parsing network (MoE-SPNet) which aims to learn predictions by considering features computed with different receptive fields (experts) and adaptively aggregate these predictions (gating network) to produce final semantic segmentation masks.  Our network is built on DeepLab-ASPP \cite{chen2016deeplab} which exploits different receptive fields for scene parsing. 

\begin{figure*}[ht!]

\begin{center}
\begin{subfigure}{0.5\textwidth}
  \begin{center}
  \includegraphics[scale=0.4]{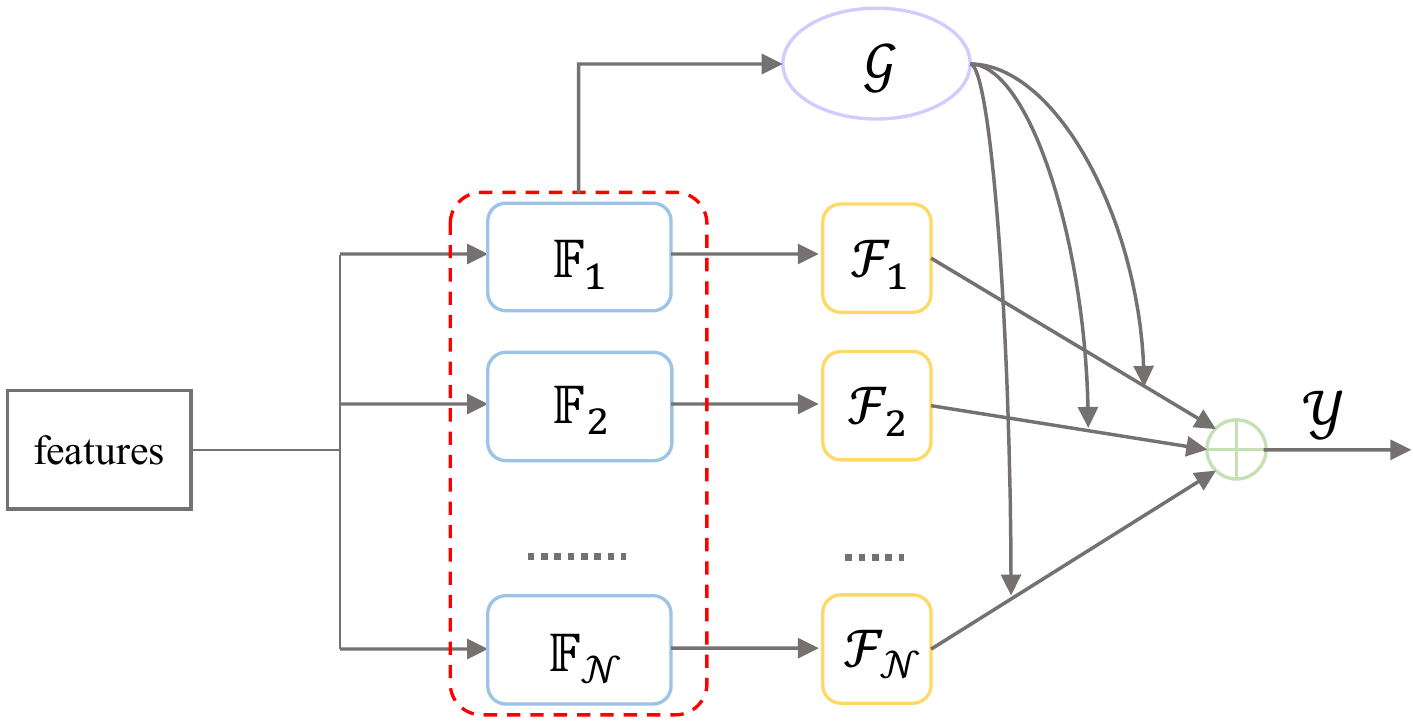}
  \end{center}
  \end{subfigure}%
\begin{subfigure}{0.5\textwidth}
  \begin{center}
  \includegraphics[scale=0.4]{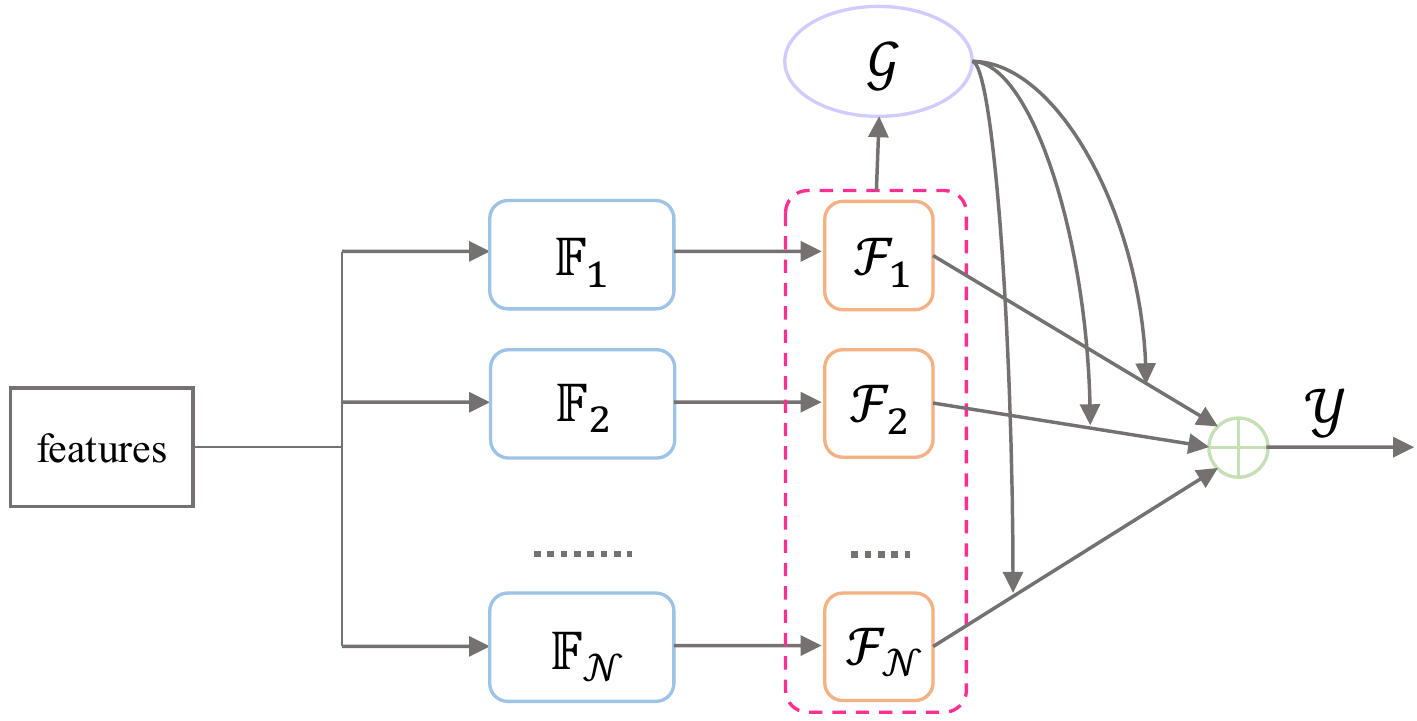}
  \end{center}
\end{subfigure}%

\caption{\small{\textbf{Variants of MoE.} Left: Learning the gating network from high-level features $\mathbb{F}_i$ of each expert.  Right: Learning the gating network from the predictions $\mathcal{F}_i$ of each expert.}}
\label{fig:MoE1}
\end{center}
\end{figure*}

Each expert in MoE-SPNet targets at learning a parsing mask from a specific receptive field. In particular, an expert adopts a dilated convolutional layer with a specific dilation rate ($e_{i}^{1}$) to obtain local structural and contextual information from features computed with a specific receptive field on top of the \texttt{pool5} layer.  Followed by two additional convolutional layers with the filter size of $1 \times 1$ ($e_{i}^{2}$ and $e_{i}^{3}$), each expert can learn a richer representation (denote as $\mathbb{F}_{i}$) of the input scene, and produce a solution (denote as $\mathcal{F}_{i}$) for the parsing task. Thus, with different dilation rates, the network can obtain some experts corresponding to different parsing solutions. \textcolor{black}{Specifically, each experts are supervised by the ground-truth parsing via \emph{softmax regression}. Thus, each channel of $\mathcal{F}_{i}$ corresponds to the probability of belonging to a category.}

\begin{figure}[ht!]

\begin{center}
\begin{subfigure}{1.0\textwidth}
  \begin{center}
  \includegraphics[scale=0.4]{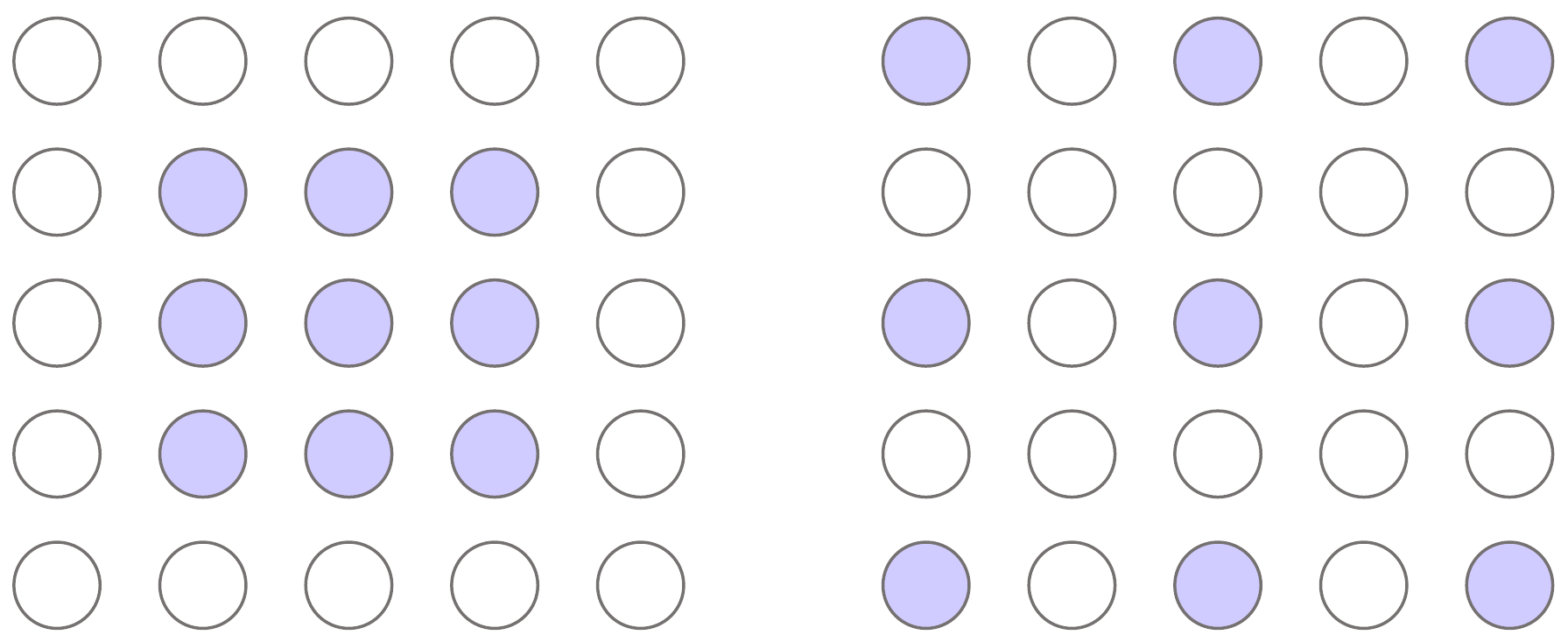}
  \end{center}
  \end{subfigure}%

\caption{\small{\textbf{Dilation.} Convolutional layers with the kernel size of 3. Left: standard convolutional layer. Right: dilated convolutional layer with the dilation rate of 2.}}
\label{fig:dilation}
\end{center}
\end{figure}

As shown in Fig. \ref{fig:MoE-spnet}, the gating network in our MoE-SPNet is different from the standard gating network that learns weights from the input features to combine a series of classifiers. We learn the gating network from the segmentation maps generated by different experts instead of the same input features fed to the experts. Fig. \ref{fig:MoE1} shows two variants of the standard MoE: one uses the high-level features $\mathbb{F}_i$ and the other uses the predictions $\mathcal{F}_i$ for the gating network. These two variants are supposed to perform better than the standard MoE, because of adoption of higher-level representations for the gating network. Since prediction maps are directly supervised using ground-truth segmentation maps, they contain the richest semantic information and are most suitable for training the gating network. Another advantage of using predictions $\mathbf{F}_i$ to train the gating network is that the number of gating network parameters can be reduced, leading to lower computational and memory cost.

To train the gating network, we concatenate $\mathcal{F}_{1}$ to $\mathcal{F}_{\mathcal{N}}$ ($\mathcal{N}$ is the number of experts), denoted as $\mathcal{F}$, and learn a non-linear function via two convolutional layers ($g_{1}'$ and $g_{2}'$) from these features to the corresponding gating features, denoted as $\mathcal{G}$, which follows the form: 
\begin{equation}
\begin{split}
\mathcal{G} = (\mathcal{F} * \mathcal{K}_{g_{1}'}) *  \mathcal{K}_{g_{2}'},
\end{split}
\label{eq:eq0}
\end{equation}
where $\mathcal{K}_{g_{1}'}$ and $ \mathcal{K}_{g_{2}'}$ are kernels of the convolutional layers $g_{1}'$ and $g_{2}'$ with the kernel size of $3 \times 3$ and $1 \times 1$ respectively,  ``$*$" is the convolution operator, and the gating features set $\mathcal{G}$ in this paper consists of $\mathcal{G}_{1}$ to $\mathcal{G}_{\mathcal{N}}$. Followed by a normalisation process, the weight located at $(i, j)$  for expert $l$ can be calculated as:
\begin{equation}
\begin{split}
w_{l}(i, j) = \frac{e^{\mathcal{G}_{l}(i, j)}} {\sum_{k=1}^{\mathcal{N}}e^{\mathcal{G}_{k}(i, j)}}.
\end{split}
\label{eq:eq00}
\end{equation}
After obtaining the weight maps in the gating networks, each channel $\mathcal{F}_{i}$ is multiplied by the corresponding weight map $\mathcal{W}_{i}$, and the aggregated output can be calculated as:
\begin{equation}
\begin{split}
\mathcal{A} = \sum_{i=1}^{\mathcal{N}} \mathcal{F}_{i} \otimes \mathcal{W}_{i},
\end{split}
\label{eq:eq000}
\end{equation}
where ``$\otimes$" denotes element-wise product in each channel.


We train MoE-SPNet using the cost function consisting of  a cooperation encouraged error term and a weakened competition encouraged error term:
 \begin{equation}
\begin{split}
\mathcal{L} = \Phi(\mathcal{Y}, \mathcal{A}) + \sum_{i=1}^{N}\Phi(\mathcal{Y}, \mathcal{F}_{i} \otimes \mathcal{W}_{i}),
\end{split}
\label{eq:eq0000}
\end{equation}
where the ``$\Phi$" represents the multinomial logistic regression error. Note that, all of the experts in our parsing network are addressing the single occasion rather than different occasions, thus the competition between these experts should not be strong. As a result, we ignore the gating factors in the typical competition encouraged error term.

\subsection{FCN-AHFA}
\label{ahfa}

\begin{figure}[ht!]

\begin{center}
\begin{subfigure}{1.0\textwidth}
  \begin{center}
  \includegraphics[scale=0.5]{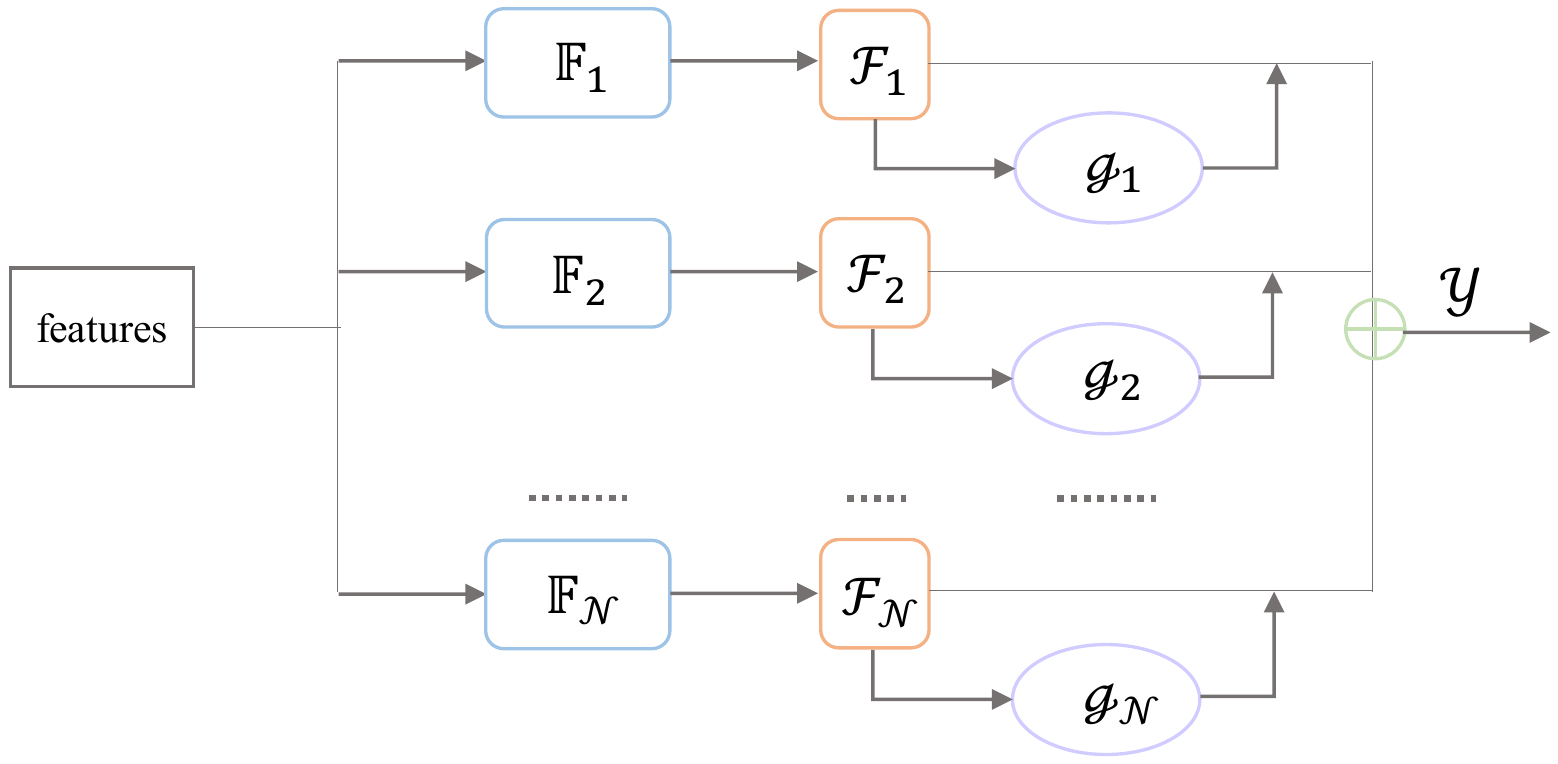}
  \end{center}
  \end{subfigure}%

\caption{\small{\textbf{AHFA:} An illustration of the proposed adaptively hierarchical features aggregation(AHFA) technique, which is another variant of mixture of experts.}}
\label{fig:MoE-ahfa}
\end{center}
\end{figure}

We investigate how to incorporate the mechanism of MoE into another popular parsing network architecture with stage-wise fusions of features from different layers. We hypothesize that the gating map for each expert can be directly learned from the expert itself and propose an adaptive hierarchical feature aggregation (AHFA) mechanism which is a variant of the proposed MoE in Sec.~\ref{sec:MoE-spnet} by assuming sparse connections in the gating network. We take the typical parsing network FCN-8s \cite{long2015fully} as an example to demonstrate the effectiveness of AHFA.

To take advantage of contextual information, FCN-8s produces a finer 16x-prediction with 16 pixel stride (16x) by adding a $1 \times 1$ convolutional layer on top of the \texttt{pool4} layer. The 32x-prediction is then upsampled to the same size of the 16x-prediction via a learnable deconvolutional layer, and then summed up with the 16x-prediction to accomplish one stage of combination. Finally,  the above combined prediction is further aggregated with higher resolution (8x) features by applying the same strategy. The final prediction with stride 8 is upsampled back to the input image resolution.

\begin{figure*}[ht!]

\begin{center}
\begin{subfigure}{1.0\textwidth}
  \begin{center}
  \includegraphics[width=0.99\textwidth]{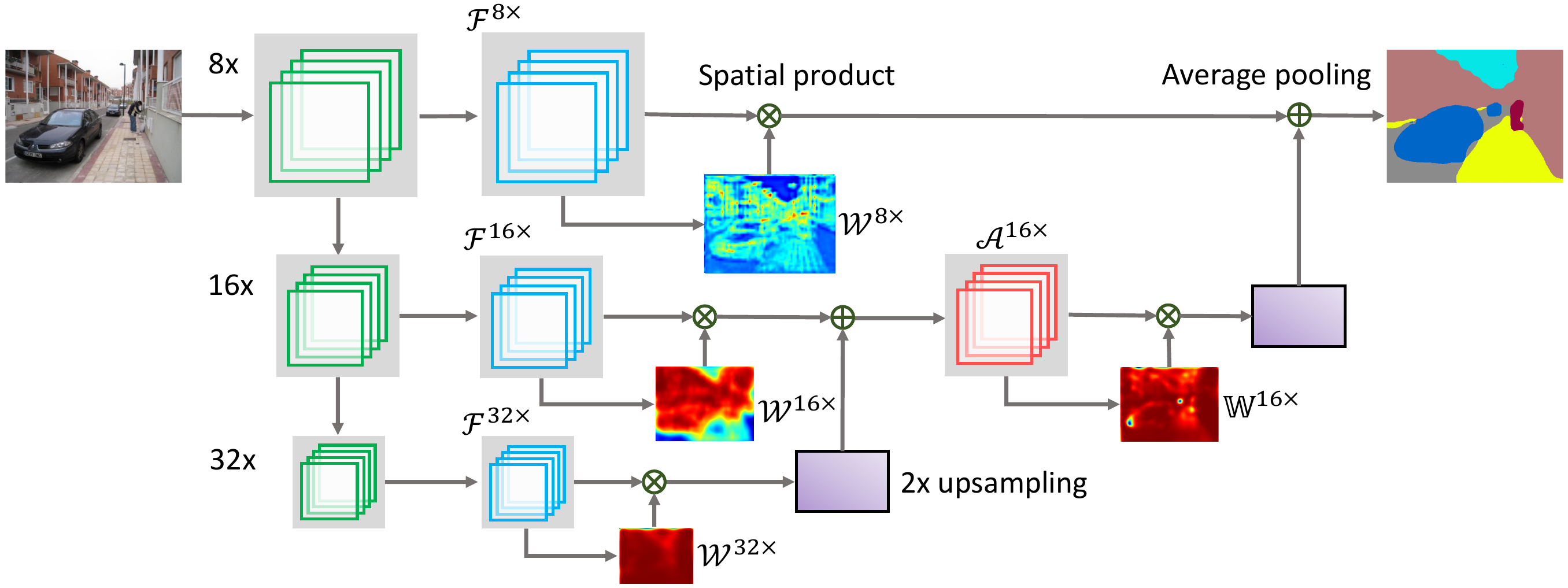}
    \end{center}
\end{subfigure}%
\caption{\small{\textbf{AHFA for skip-nets.} The feature maps are fused by stage-wise combination in skip-nets. In each combination stage, we learn a soft weight map for each level of features followed by a weighted pooling step over adjacent levels.}}
\label{fig:fcn_ahfa}
\end{center}
\end{figure*}

Now we describe the details of AHFA in FCN-8s to adaptively merge hierarchical features (32x, 16x, and 8x), resulting in a modified model which we call FCN-AHFA. An illustration of FCN-AHFA is shown in Fig. \ref{fig:fcn_ahfa}. In the first stage, on top of the 32x-prediction, denoted as $\mathcal{F}^{\text{32x}}\in\mathbb{R}^{H\times W\times C}$, we add a convolutional layer with the kernel size of $3 \times 3$ and the stride of $1$ followed by a sigmoid layer to produce a dense probabilistic weight map $\mathcal{W}^{\text{32x}}\in\mathbb{R}^{H\times W}$. Here, $H$, $W$, and $C$ denote the height, width, and the number of channels of the 32x feature maps, respectively. Then the weight located at $(i, j)$ in $\mathcal{W}^{\text{32x}}$ can be calculated as:
\begin{equation}
\begin{split}
w_{(i,j)}^{\text{32x}} = \frac{1}{1 + e^{- \sum_{c=1}^{C}(f_{c}^{\text{32x}} * k_{c}^{\text{32x}})(i, j)}},
\end{split}
\label{eq:eq1}
\end{equation}
where $f_{c}^{\text{32x}}$ represents the $c$-th channel of $\mathcal{F}^{\text{32x}}$, $k_{c}^{\text{32x}}$ is the corresponding convolutional kernel, and ``$*$" is the convolution operator. The weight  function in Eq. (\ref{eq:eq1}) can be made more complex by introducing more convolutional and activation layers. However, we have observed from the experimental results that learning more complex weight functions only slightly improves the performance.
After obtaining the weight map, each channel of $\mathcal{F}^{\text{32x}}$ is multiplied by $\mathcal{W}^{\text{32x}}$, resulting in the weighted features $\mathcal{H}^{\text{32x}}\in\mathbb{R}^{H\times W\times C}$ of which each channel is:
\begin{equation}
\begin{split}
h_{c}^{\text{32x}} = \mathcal{W}^{\text{32x}} \otimes f_{c}^{\text{32x}},
 \end{split}
\label{eq:eq2}
\end{equation}
where $\otimes$ represents Hadamard product or entrywise product. Likewise, we reweight the 16x-prediction $\mathcal{F}^{\text{16x}}$ by the learned weight $\mathcal{W}^{\text{16x}}$ to obtain $\mathcal{H}^{\text{16x}}\in\mathbb{R}^{2H\times 2W\times C}$. At the final step of this stage, $\mathcal{H}^{\text{32x}}$ is upsampled to have the same size of the $\mathcal{H}^{\text{16x}}$ and linearly combined with it to produce the 16x aggregated feature:
\begin{equation}
\begin{split}
\mathcal{A}^{\text{16x}} = \mathcal{H}^{\text{16x}} \oplus (\mathcal{H}^{\text{32x}})^{\uparrow},
 \end{split}
\label{eq:eq3}
\end{equation}
where $(\bullet)^{\uparrow}$ is a 2x upsampling operation via bilinear interpolation and $\oplus$ denotes the summing operation in each spatial location.

The aggregation strategy for the second stage is similar to that used in the first stage but is applied on $\mathcal{A}^{\text{16x}}$ and $\mathcal{F}^{\text{8x}}$. Hence, the the $c$-th channel of $\mathcal{A}^{\text{8x}}$ can be calculated as:
\begin{equation}
\begin{split}
a_{c}^{\text{8x}} =  (\mathcal{W}^{\text{8x}} \otimes f_{c}^{\text{8x}})\oplus (\mathbb{W}^{\text{16x}} \otimes a_{c}^{\text{16x}})^{\uparrow},
 \end{split}
\label{eq:eq4}
\end{equation}
where $\mathcal{W}^{\text{8x}}$ and $\mathbb{W}^{\text{16x}}$ are the learned probabilistic weight maps for $\mathcal{F}^{\text{8x}}$ and $\mathcal{A}^{\text{16x}}$, repectively. 
\newline

\noindent\textbf{Remark} The fixed-size filters ($3 \times 3$) used for learning the weight maps are actually adaptive to the size of semantic areas in the input image, because the higher-layer feature maps have smaller size. For example, the spatial areas corresponding to the original image considered by $k_{c}^{\text{32x}}$ are four times larger than that considered by $k_{c}^{\text{16x}}$.  Also, the weight map of a layer is learned only from the feature maps in that layer. This is different from existing mixture-of-experts \cite{jordan1994hierarchical} or the attention models \cite{CY2016Attention} which usually learn the weights from the concatenation of features maps from all layers. Our method simplifies the weight learning network based on the observation that the feature maps in one layer already contain rich information about the corresponding weight map. Finally, with the learned weight maps,  different levels of features can be aggregated adaptively by considering the relative spatial-semantic tradeoff at each spatial location.

\section{Experiments}
To demonstrate the effectiveness of the proposed MoE-SPNet and FCN-AHFA methods, we compare our methods with the existing methods on two challenging datasets, \ie PASCAL VOC 2012 \cite{everingham2015pascal} and SceneParse150 \cite{zhou2016semantic}. We first describe the experimental settings including evaluation protocols and detailed implementations, and then report the experimental results with discussions.

\subsection{Experimental Setting}
\noindent\textbf{Evaluation Metrics} Four common metrics for scene parsing are used in our experiments, \ie pixel accuracy, mean accuracy, mean IoU, and weighted IoU. Pixel accuracy indicates the proportion of correctly classified pixels. Mean accuracy indicates the average of the proportion of correctly classified pixels for all classes. IoU indicates the average intersection-over-union between the predicted and ground-truth pixels over all classes. Weighted IoU indicates the IoU weighted by total pixel ratio of each class. Let $L$ be the number of classes of interest, $l_{ij}$ represents the number of pixels belonging to class $i$ predicted as class $j$, and ${N} = \sum_{i=1}^{L}\sum_{j=1}^{L}l_{ij}$ is the number of pixels. The four metrics are computed as follows:

\begin{itemize}
\item[$\bullet$] Pixel Acc. : $\frac{1}{{N}}\sum_{i=1}^{L}l_{ii}$
\item[$\bullet$] Mean Acc. : $\frac{1}{L} \sum_{i=1}^{L}\frac{l_{ii}}{\sum_{j=1}^{L}l_{ij}}$
\item[$\bullet$] Mean IoU : $\frac{1}{L} \sum_{i=1}^{L}\frac{l_{ii}}{- l_{ii} + \sum_{j=1}^{L}(l_{ij} + l_{ji})}$
\item[$\bullet$] Weighted IoU : $\frac{1}{N}\sum_{i=1}^{L}\frac{l_{ii}\sum_{j=1}^{L}l_{ij}}{- l_{ii} + \sum_{j=1}^{L}(l_{ij} + l_{ji})}$
\end{itemize}

It should be noted that pixel accuracy is biased to reflect the ``stuff" categories such as grass and sky as they occupy more pixels. Instead, IoU is a more accurate measure of the classification performance on ``things" categories such as person and car. 
\newline

\noindent\textbf{Implementation} Since the proposed methods rely on semantic predictions in each level, our framework are trained in two stages. In the first stage, we train the basic network without MoE to produce hierarchical features containing  semantic information. In the second stage, we add the gating network of MoE to the pre-trained baseline network and fine-tune the whole parsing network in an end-to-end fashion. For fair comparison, we also fine-tune the baseline network with the same iterations. We initialise the base convolutional architecture via the pre-trained VGG16, ResNet-50, and ResNet-101 \cite{he2016deep} classification models on ILSVRC \cite{imagenet_cvpr09}. The fine-tuning stage follows a polynomial decay with the power of 0.9, the momentum of 0.9, and the weight decay of 0.0005. We implement our networks based on \emph{Caffe} \cite{jia2014caffe}, and train them using 4 TITAN X GPUs with 12GB of memory per GPU. The batch size is set to 8 in all the experiments.
\newline

\noindent\textbf{Data Augmentation} Data augmentation techniques are used when training the parsing networks, which can be summarised as follows: 
1). The training images are resized by the scaling factors: 0.5, 0.75, 1.0, 1.25, 1.5. 2). We randomly flip the training images horizontally. 3). The input samples of the models are randomly cropped from the training images with a fixed size.

\subsection{Benchmark Performance}

\subsubsection{PASCAL VOC 2012}

\begin{figure*}[ht!]

\begin{center}
\begin{subfigure}{1.0\textwidth}
  \begin{center}
  \includegraphics[width=0.99\textwidth]{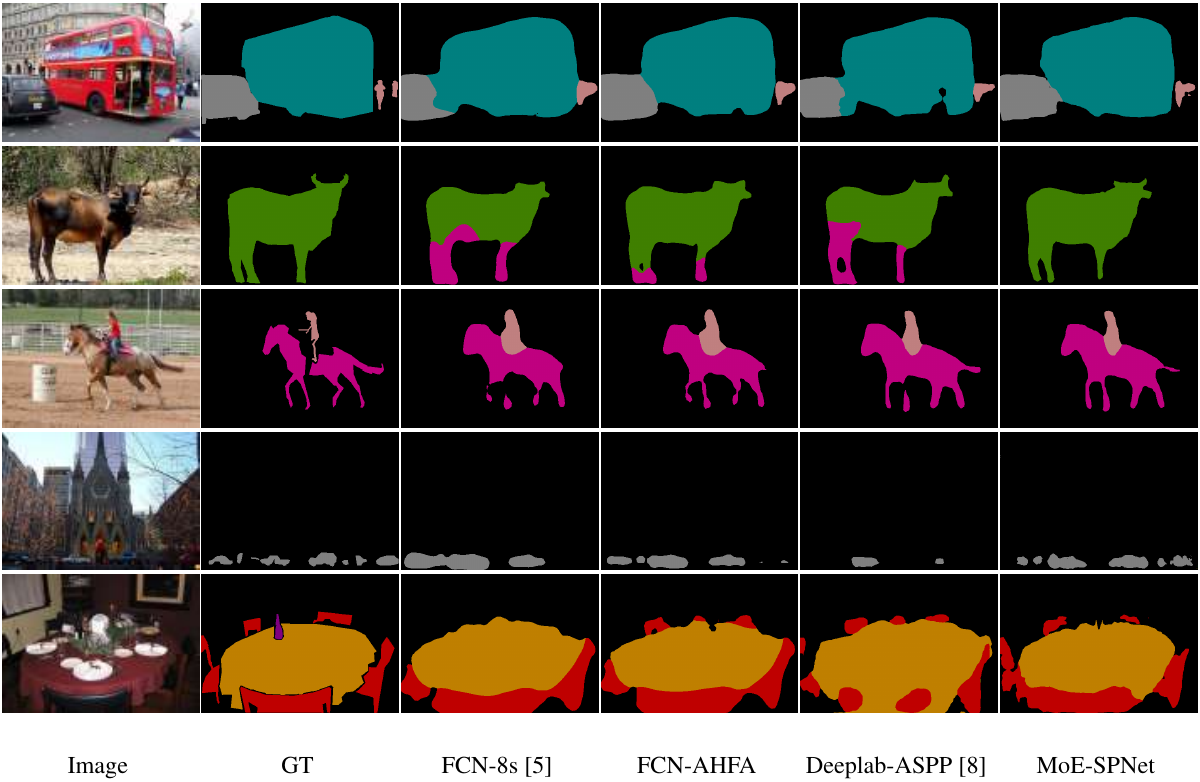}
    \end{center}
\end{subfigure}%
\caption{\small{\textbf{PASCAL VOC 2012 results.}  A comparison of proposed MoE based parsing networks, \ie FCN-AHFA, MoE-SPNet, with their baseline models, \ie FCN-8s, DeepLab-ASPP. (Best view in colour.)}}
\label{fig:voc}
\end{center}
\end{figure*}

PASCAL VOC 2012 \cite{pascal-voc-2012}, which consists of 20 common object categories and one background category, is a well-known benchmark for semantic segmentation. The images contained in this dataset are split into three parts, including 1464 training images, 1449 validation images, and 1456 test images. Following \cite{hariharan2011semantic}, the training data with ground-truth segmentation masks are augmented to 10,582 images using the extra annotated images for VOC 2012. Since PASCAL VOC 2012 is an object-level segmentation benchmark, and each image in this dataset follows a simple foreground/background form. We only adopt mean IoU, which is a stricter and more convincing metric for scene parsing, to evaluate different methods following previous works.

 \setlength\tabcolsep{0.9pt}
\begin{table}[h!]
\scriptsize
\centering
\begin{tabular}{ l || c || c | c | c | c | c | c | c | c | c | c | c | c | c | c | c | c | c | c | c | c }

& \rotatebox{90}{\textbf{mean}}  &\rotatebox{90}{areo} & \rotatebox{90}{bike} & \rotatebox{90}{bird} & \rotatebox{90}{boat} & \rotatebox{90}{bottle} & \rotatebox{90}{bus} & \rotatebox{90}{car} & \rotatebox{90}{cat} & \rotatebox{90}{chair} & \rotatebox{90}{cow} & \rotatebox{90}{table} & \rotatebox{90}{dog} & \rotatebox{90}{horse} & \rotatebox{90}{mbike} & \rotatebox{90}{person} & \rotatebox{90}{plant} & \rotatebox{90}{sheep} & \rotatebox{90}{sofa} & \rotatebox{90}{train} & \rotatebox{90}{tv} \\
\hline\hline
\multicolumn{22}{  c  }{  VGG $+$ PASCAL VOC } \\
\hline
\tiny{SegNet \cite{badrinarayanan2017segnet}}  & 59.9 &  73.6 & 37.6 & 62.0 & 46.8 & 58.6 & 79.1 & 70.1 & 65.4 & 23.6 & 60.4 & 45.6 & 61.8 & 63.5 & 75.3 & 74.9 & 42.6 & 63.7 & 42.5 & 67.8 & 52.7 \\
\tiny{FCN-8s \cite{long2015fully}}  & 62.2 &  76.8 & 34.2 & 68.9 & 49.4 & 60.3 & 75.3 & 74.7 & 77.6 & 21.4 & 62.5 & 46.8 & 71.8 & 63.9 & 76.5 & 73.9 & 45.2 & 72.4 & 37.4 & 70.9 & 55.1 \\
\tiny{Hypercolumn \cite{hariharan2015hypercolumns}}  & 62.6 &  68.7 & 33.5 & 69.8 & 51.3 & 70.2 & 81.1 & 71.9 & 74.9 & 23.9 & 60.6 & 46.9 & 72.1 & 68.3 & 74.5 & 72.9 & 52.6 & 64.4 & 45.4 & 64.9 & 57.4 \\
\tiny{Zoom-out \cite{mostajabi2015feedforward}} & 69.6 &  85.6 & 37.3 & 83.2 & 62.5 & 66.0 & 85.1 & 80.7 & 84.9 & 27.2 & 73.2 & 57.5 & 78.1 & 79.2 & 81.1 & 77.1 & 53.6 & 74.0 & 49.2 & 71.7 & 63.3 \\
\tiny{EdgeNet \cite{chen2016semantic}} & 71.2 &  83.6 & 35.8 & 82.4 & 63.1 & 68.9 & 86.2 & 79.6 & 84.7 & 31.8 & 74.2 & 61.1 & 79.6 & 76.6 & 83.2 & 80.9 & 58.3 & 82.6 & 49.1 & 74.8 & 65.1 \\
\tiny{Attention \cite{CY2016Attention}} & 71.5 &  86.0 & 38.8 & 78.2 & 63.1 & 70.2 & 89.6 & 84.1 & 82.9 & 29.4 & 75.2 & 58.7 & 79.3 & 78.4 & 83.9 & 80.3 & 53.5 & 82.6 & 51.5 & 79.2 & 64.2 \\
\tiny{DeepLab-Large \cite{chen2014semantic}}  & 71.6 &  84.4 & \textbf{54.5} & 81.5 & 63.6 & 65.9 & 85.1 & 79.1 & 83.4 & 30.7 & 74.1 & 59.8 & 79.0 & 76.1 & 83.2 & 80.8 & 59.7 & 82.2 & 50.4 & 73.1 & 63.7 \\  
\tiny{CRFasRNN \cite{zheng2015conditional}} & 72.0 &  87.5 & 39.0 & 79.7 & 64.2 & 68.3 & 87.6 & 80.8 & 84.4 & 30.4 & 78.2 & 60.4 & 80.5 & 77.8 & 83.1 & 80.6 & 59.5 & 82.8 & 47.8 & 78.3 & 67.1 \\ 
\tiny{DeconvNet \cite{noh2015learning}}  & 72.5 &  89.9 & 39.3 & 79.7 & 63.9 & 68.2 & 87.4 & 81.2 & 86.1 & 28.5 & 77.0 & 62.0 & 79.0 & 80.3 & 83.6 & 80.2 & 58.8 & 83.4 & 54.3 & 80.7 & 65.0 \\ 
\tiny{DPN \cite{liu2015semantic}}    & 74.1 &  87.7 & 59.4 & 78.4 & 64.9 & 70.3 & 89.3 & 83.5 & 86.1 & 31.7 & 79.9 & 62.6 & 81.9 & 80.0 & 83.5 & 82.3 & 60.5 & 83.2 & 53.4 & 77.9 & 65.0 \\
\tiny{Cont-CNN-CRF \cite{lin2015efficient}} & 75.3 &  90.6 & 37.6 & 80.0 & \textbf{67.8} & \textbf{74.4} & 92.0 & 85.2 & 86.2 & \textbf{39.1} & 81.2 & 58.9 & 83.8 & 83.9 & 84.3 & \textbf{84.8} & 62.1 & 83.2 & \textbf{58.2} & 80.8 & \textbf{72.3} \\
\hline
\tiny{MoE-SPNet}  & 74.7 &  90.1 & 38.6 & 79.7 & 63.4 & 69.9 & 90.9 & 86.4 & \textbf{89.1} & 32.2 & \textbf{82.7} & 62.6 & \textbf{84.9} & 83.3 & \textbf{85.7} & 82.7 & \textbf{63.9} & 84.2 & 56.6 & 79.3 & 67.6 \\
\tiny{FCN-AHFA}  & 70.6 &  82.6 & 37.2 & 80.9 & 58.0 & 67.7 & 86.4 & 84.6 & 84.5 & 30.2 & 76.6 & 50.3 & 78.7 & 79.1 & 83.4 & 80.3 & 59.3 & 78.5 & 48.5 & 80.5 & 61.9 \\

\hline\hline
\multicolumn{22}{  c  }{  VGG $+$ PASCAL VOC $+$ COCO} \\
\hline
\tiny{EdgeNet \cite{chen2016semantic}} & 73.6 &  88.3 & 37.0 & 89.8 & 63.6 & 70.3 & 87.3 & 82.0 & 87.6 & 31.1 & 79.0 & 61.9 & 81.6 & 80.4 & 84.5 & 83.3 & 58.4 & 86.1 & 55.9 & 78.2 & 65.4 \\
\tiny{CRFasRNN \cite{zheng2015conditional}}& 74.7 &  90.4 & 55.3 & 88.7 & 68.4 & 69.8 & 88.3 & 82.4 & 85.1 & 32.6 & 78.5 & 64.4 & 79.6 & 81.9 & 86.4 & 81.8 & 58.6 & 82.4 & 53.5 & 77.4 & 70.1 \\
\tiny{BoxSup \cite{dai2015boxsup}} & 75.2 &  89.8 & 38.0 & 89.2 & 68.9 & 68.0 & 89.6 & 83.0 & 87.7 & 34.4 & 83.6 & 67.1 & 81.5 & 83.7 & 85.2 & 83.5 & 58.6 & 84.9 & 55.8 & 81.2 & 70.7 \\
\tiny{SBound \cite{kokkinos2015pushing}}& 75.7 &  90.3 & 37.9 & 89.6 & 67.8 & 74.6 & 89.3 & 84.1 & 89.1 & 35.8 & 83.6 & 66.2 & 82.9 & 81.7 & 85.6 & 84.6 & 60.3 & 84.8 & 60.7 & 78.3 & 68.3 \\
\tiny{Attention \cite{CY2016Attention}} & 76.3 &  93.2 & 41.7 & 88.0 & 61.7 & 74.9 & 92.9 & 84.5 & 90.4 & 33.0 & 82.8 & 63.2 & 84.5 & 85.0 & 87.2 & 85.7 & 60.5 & 87.7 & 57.8 & 84.3 & 68.2 \\
\tiny{DPN \cite{liu2015semantic}} & 77.5 & 89.0 & \textbf{61.8} & 87.7 & 66.8 & 74.7 & 91.2 & 84.3 & 87.6 & 36.5 & 86.3 & 66.1 & 84.4 & 87.8 & 85.6 & 85.4 & 63.6 & 87.3 & 61.3 & 79.4 & 66.4 \\
\tiny{Cont-CNN-CRF \cite{lin2015efficient}} & 77.8 &  \textbf{94.1} & 40.4 & 83.6 & 67.3 & 75.6 & 93.4 & 84.4 & 88.7 & 41.6 & 86.4 & 63.3 & 85.5 & 89.3 & 85.6 & 86.0 & \textbf{67.4} & \textbf{90.1} & 62.6 & 80.9 & 72.5 \\
\tiny{TVG-HO-CRF \cite{arnab2016higher}} & 77.9 & 92.5 & 59.1 & \textbf{90.3} & \textbf{70.6} & 74.4 & 92.4 & 84.1 & 88.3 & 36.8 & 85.6 & 67.1 & 85.1 & 86.9 & 88.2 & 82.6 & 62.6 & 85.0 & 56.3 & 81.9 & \textbf{72.5} \\
\tiny{Att-CRF-DT \cite{chen2016semantic}} & 76.3 &  93.2 & 41.7 & 88.0 & 61.7 & 74.9 & 92.9 & 84.5 & 90.4 & 33.0 & 82.8 & 63.2 & 84.5 & 85.0 & 87.2 & 85.7 & 60.5 & 87.7 & 57.8 & 84.3 & 68.2 \\
\hline
\tiny{MoE-SPNet} & 77.7 &  91.6 & 39.7 & 89.6 & 64.2 & \textbf{77.1} & 93.7 & \textbf{89.0} & \textbf{93.6} & 36.5 & \textbf{87.6} & 56.0 & \textbf{90.3} & \textbf{91.6} & 85.9 & 86.7 & 59.2 & 89.3 & 59.3 & \textbf{85.7} & 70.9 \\
\hline\hline
\multicolumn{22}{  c  }{  ResNet-101 $+$ PASCAL VOC $+$ COCO} \\
\hline
\tiny{Deeplab-ASPP \cite{chen2016deeplab}} & 79.7	& 92.6 & 60.4 & 91.6	 & 63.4 & 76.3 & 95.0 & 88.4 & 92.6 & 32.7 & 88.5 & 67.6 & 89.6 & 92.1 & 87.0 & 87.4 & 63.3 & 88.3	 & 60.0 & 86.8 & 74.5 \\
\tiny{LRR-CRF \cite{liu2015semantic}} & 79.3 & 92.4 & 45.1 & 94.6 & 65.2 & 75.8 & 95.1 & 89.1 & 92.3 & 39.0 & 85.7 & 70.4 & 88.6 & 89.4 & 88.6 & 86.6 & 65.8 & 86.2 & 57.4 & 85.7 & 77.3 \\
\tiny{Deep G-CRF \cite{chandra2016fast}} & 80.2 & 92.9 & 61.2 & 91.0 & 66.3	 & 77.7 & \textbf{95.3} & 88.9 & 92.4 & 33.8 & 88.4 & 69.1 & 89.8 & 92.9 & 87.7 & 87.5 & 62.6 & 89.9 & 59.2 & \textbf{87.1} & 74.2 \\
\tiny{FRRN \cite{pohlen2017full}} & 80.3	& 94.4 & 61.3 & 91.1	 & 65.7 & 76.2 & 94.5 & 88.1 & 91.9 & 35.1 & 89.2 & 70.9 & 88.6 & 92.3 & 87.9 & 87.9 & 62.9 & 89.9 & 61.7 & 86.6 & 74.6 \\
\tiny{Multi-Refine \cite{Lin:2017:RefineNet}} & 82.4 & \textbf{94.9} & 60.2 & 92.8 & \textbf{77.5} & 81.5 & 95.0 & 87.4 & 93.3 & 39.6 & \textbf{89.3} & \textbf{73.0} & \textbf{92.7} & 92.4 & 85.4 & \textbf{88.3} & \textbf{69.7} & \textbf{92.2} & 65.3 & 84.2 & \textbf{78.7} \\
\hline
\tiny{MoE-SPNet} & \textbf{82.5} &  94.1 & \textbf{63.9} & \textbf{93.8} & 72.3 & \textbf{82.1} & 95.2 & \textbf{89.8} & \textbf{94.2} & \textbf{40.1} & 88.1 & 70.3 & 90.0 & \textbf{93.9} & \textbf{90.0} & 87.2 & 67.0 & 91.3 & \textbf{67.0} & \textbf{87.1} & 78.2 \\
\hline
\end{tabular}
\caption{\small{\textbf{Results on PASCAL VOC 2012 test set.} For a fair comparison, In the bottom part of the table, we only compare results with previous works who also adopt the standard ResNet-101 as their base network. Thus, some works who modify the ResNet-101 to deeper or wider for their parsing network are not reported.}} 
\label{tab:voc-scores}
\end{table}

In Tab.~\ref{tab:voc-scores}, we report our scores on the test server in different conditions to make a comparison with previous works. Our MOE-SPNet achieves about 2.5$\%$ improvement on the baseline model Deeplab-ASPP based on ResNet, and obtains comprisable results with current state-of-the-art algorithms on different settings. Also, our FCN-AHFA significantly outperforms the most typical baseline model FCN-8s, nearly closing the gap to current state-of-the-art methods. Fig. \ref{fig:voc} gives qualitative comparison of different methods on several images.

\begin{figure}[p]

\begin{center}

\begin{subfigure}{1.0\textwidth}
  \begin{center}
  \includegraphics[width=0.99\textwidth]{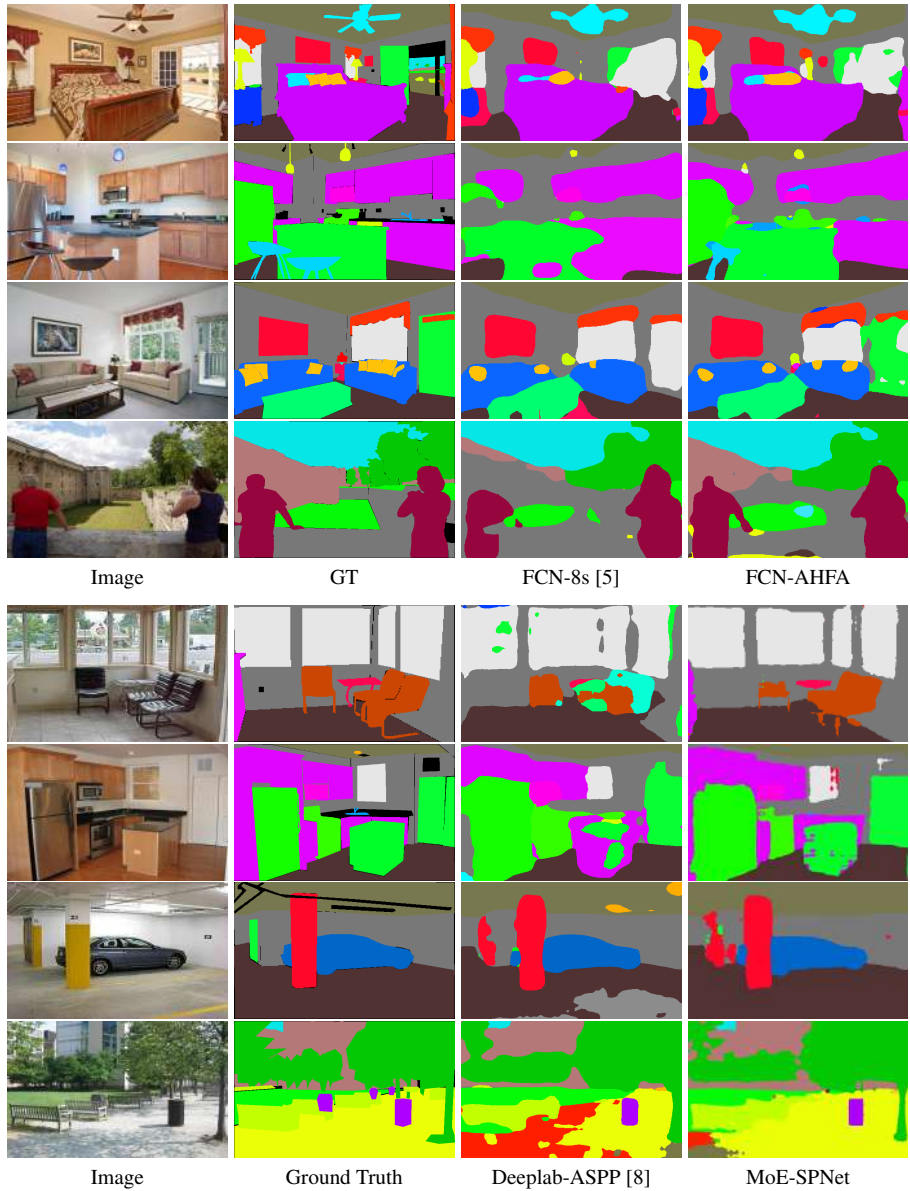}
    \end{center}
\end{subfigure}%

\caption{\small{\textbf{SceneParse150 results.} The top part shows the segmentation results of FCN-8s \cite{long2015fully} without or with our AHFA technique. The bottom part shows the segmentation results of Deeplab-ASPP \cite{chen2016deeplab} (with atrous spatial pyramid pooling) and our MOE-SPNet. (Best viewed in colour)
}}
\label{fig:ade}
\end{center}
\end{figure}

\subsubsection{SceneParse150}

\begin{figure*}[ht!]

\begin{center}

\begin{subfigure}{1.0\textwidth}
  \begin{center}
  \includegraphics[width=0.99\textwidth]{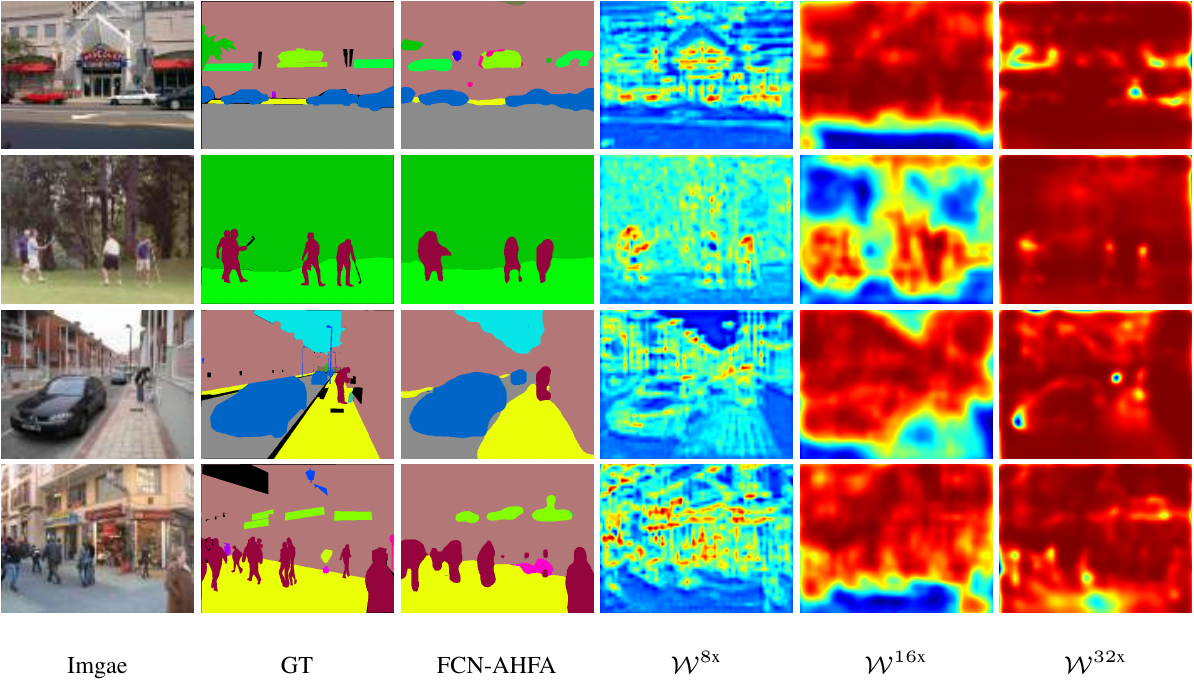}
    \end{center}
\end{subfigure}%

\caption{\small{\textbf{Weight maps on SceneParse150 by FCN-AHFA.}  Red represents high probability, and blue represents low probability. $\mathcal{W}^{8\text{x}}$, $\mathcal{W}^{16\text{x}}$, and $\mathcal{W}^{32\text{x}}$ correspond to weight maps of $\mathcal{F}^{8\text{x}}$, $\mathcal{F}^{16\text{x}}$, and $\mathcal{F}^{32\text{x}}$, respectively, as defined in section \ref{ahfa}. $\mathcal{W}^{8\text{x}}$ give high weights to the boundary positions. $\mathcal{W}^{16\text{x}}$ sets high probabilities to regions of small categories for $\mathcal{F}^{16\text{x}}$. And the $\mathcal{W}^{32\text{x}}$ activates almost all the positions, since 32x features in FCN-8s capture the global contextual information which is useful for many categories. (Best viewed in colour)}}
\label{fig:att1}
\end{center}
\end{figure*}

SceneParse150 \cite{zhou2016semantic} is a recently released large-scaled scene parsing benchmark using images from ADE20K Dataset \cite{zhou2016semantic}. The images are collected from diverse outdoor and indoor scenes involving 35 stuff categories (\eg floor, water, and sky) and 115 discrete objects (\eg person, chair, and car). We train our models on the 20,210 training images, and evaluate the performance on the 2,000 validation images. 

\begin{table}[h!]
\small
\centering
\begin{tabular}{ l || c | c | c | c }
\hline
\multirow{ 2 }{*}{Algorithm} & \multicolumn{4}{  | c  }{ Metric  } \\ \cline{2-5}
 & Pixel Acc. & Mean Acc. & Mean IoU & Weighted IoU \\
\hline\hline
\multicolumn{5}{  c  }{  VGG  } \\
\hline
Cascade-DilatedNet \cite{zhou2016semantic} & 74.52 $\%$ & 45.38 $\%$ & 0.3496 & 0.6108 \\
\hline
FCN-8s \cite{long2015fully} & 71.56 $\%$ & 40.50 $\%$ & 0.2948 & 0.5755 \\
FCN-AHFA  & \textbf{73.59} $\%$ & \textbf{43.51} $\%$ & \textbf{0.3128} & \textbf{0.6009} \\
\hline
DeepLab-ASPP \cite{chen2016deeplab} & 74.88 $\%$ & 46.17 $\%$ & 0.3303 & 0.6167 \\
MoE-SPNet& \textbf{75.50} $\%$ & \textbf{47.33} $\%$ & \textbf{0.3435} & \textbf{0.6242} \\
\hline\hline
\multicolumn{5}{  c  }{  ResNet  } \\
\hline
FCN-16s \cite{long2015fully}  & 75.52 $\%$ & 44.13 $\%$ & 0.3475 & 0.6246 \\
FCN-AHFA & \textbf{76.04} $\%$ & \textbf{45.40} $\%$ & \textbf{0.3549} & \textbf{0.6286} \\
\hline
Deeplab-ASPP \cite{chen2016deeplab}  & 77.31 $\%$ & 47.69 $\%$ & 0.3675 & 0.6354 \\
MoE-SPNet & \textbf{78.02} $\%$ & \textbf{48.02} $\%$ & \textbf{0.3789} & \textbf{0.6426} \\
\hline
\end{tabular}
\caption{\small{\textbf{Results on SceneParse150 validation set.} We add AHFA and MOE to two kinds of parsing networks, including FCN-16/8s and Deeplab-ASPP,  respectively. The comparison with baseline models demonstrates the effectiveness of our attention strategies. 
}} 
\label{tab:ade}
\end{table}

From the analyzation of different variants of MoE-SPNet, we choose MoE-SPNet, which contains least parameters but has promise performance, as our network architecture to demonstrate the effectiveness of our MoE based parsing algorithm. Also, following the same setting as the previous section, we employ the AHFA, which is the key component of MoE-SPNet, to the FCN.
Note that, when building on ResNet, we add an extra convolutional layer with kernel size of $7 \times 7$ on top of ResNet to make the learned representation more complicated for FCN. We use FCN-16s as the baseline model, as we observed from the experiments that it outperforms FCN-8s on this dataset.
Due to limited GPU memory, we use 50-layer ResNet for FCN-16s and FCN-AHFA and use ResNet101 to build two models based on DeepLab-ASPP and MoE-SPNet. 
Following the benchmark providers, we take the mean of \emph{Pixel Acc.} and \emph{Mean IoU} as the evaluation score. 

The evaluation results are shown in Tab. \ref{tab:ade}. Sampled qualitative results are shown in Fig. \ref{fig:ade}. It can be seen that the AHFA-based methods outperform the baseline methods in terms of both pixel accuracy and IoU. Our VGG16-based FCN-AHFA yields a score of 52.4\%, bringing 1.9\% improvement over the VGG16-based FCN-8s (50.5\%); and ResNet-based FCN-AHFA obtains a score of 55.8\%, outperforming ResNet-based FCN-16s (55.1\%) by 0.7\%.  Also, some selected learned weight maps in Fig. ~\ref{fig:att1} furtherly demonstrate the effectiveness of our attention strategy.

Also,  VGG-based MoE-SPNet has a score of 54.9\%, yielding  0.9\% improvement over the baseline method DeepLab-ASPP. Also, ResNet-based MoE-SPNet achieves 58.0\%, which is 1\% higher than performance of ResNet-based DeepLab-ASPP. It should be noted that obtaining 1\% overall improvement on this dataset containing 150 classes is considered as significant. Especially, the MoE-SPNet method outperforms the Cascade-DilatedNet \cite{zhou2016semantic} which segments stuff, objects, and object parts via a complicated cascade structure.

\subsection{Ablation Studies on PASCAL VOC}
We run some experiments to analyze our MOE and AHFA based networks, and discuss them in detail here.

\subsubsection{Comparison with Baseline}

 \setlength\tabcolsep{3.0pt}
\begin{table}[h!]
\normalsize
\centering
\begin{tabular}{ l || c | c }
\hline
\multirow{ 2 }{*}{Algorithm} & \multicolumn{2}{c}{ Mean IoU (\%) } \\ \cline{2-3}
 & val & test \\
\hline\hline
FCN-8s \cite{long2015fully} & 68.4  & 62.2\\
FCN-AHFA  & 70.5 & 70.6  \\
\hline
DeepLab-ASPP \cite{chen2016deeplab} & 66.3 & 72.6  \\
MoE-SPNet& 70.4 & 74.7  \\
\hline
\end{tabular}
\caption{\small{\textbf{Comparison with Baseline} This table reports the mean IoU on PASCAL VOC 2012 on val/test set. The FCN-8s and Deeplab-ASPP are two baseline networks.}} 
\label{tab:voc-validation}
\end{table}

To demonstrate the effectiveness of our MOE-SPNet, we compare VGG16-based MoE-SPNet with the baseline parsing network Deeplab-ASPP \cite{chen2016deeplab} on both the validation set and the test set. We also apply AHFA to the popular stage-wise parsing network FCN-8s to demonstrate the wide applicability of proposed AHFA based parsing strategy. As shown in Tab.~\ref{tab:voc-validation}, our methods consistently outperform the counterpart baseline networks. In particular, our MoE-SPNet obtains 2.1\% improvement in terms of \emph{Mean IoU} compared with the baseline Deeplab-ASPP \cite{chen2016deeplab} on both sets. Significantly, employing our AHFA method to FCN-8s results in an \emph{Mean IoU} of 70.4\%, which not only outperforms FCN-8s (66.3\%) by 4.1\%, but also achieves comparable performance with the state-of-the-art algorithms. It should be noted that, this comparison does not adopt the extra performance boosting techniques like CRF post-processing, pre-training on MS COCO, or multi-scale inputs. 

\subsubsection{Variants of MOE-SPNet}

 \setlength\tabcolsep{0.9pt}
\begin{table}[h!]
\scriptsize
\centering
\begin{tabular}{ l || c || c | c | c | c | c | c | c | c | c | c | c | c | c | c | c | c | c | c | c | c }

& \rotatebox{90}{mean}  &\rotatebox{90}{areo} & \rotatebox{90}{bike} & \rotatebox{90}{bird} & \rotatebox{90}{boat} & \rotatebox{90}{bottle} & \rotatebox{90}{bus} & \rotatebox{90}{car} & \rotatebox{90}{cat} & \rotatebox{90}{chair} & \rotatebox{90}{cow} & \rotatebox{90}{table} & \rotatebox{90}{dog} & \rotatebox{90}{horse} & \rotatebox{90}{mbike} & \rotatebox{90}{person} & \rotatebox{90}{plant} & \rotatebox{90}{sheep} & \rotatebox{90}{sofa} & \rotatebox{90}{train} & \rotatebox{90}{tv} \\
\hline\hline
Baseline & 72.6 & 88.3 & 37.0 & 89.8 & 63.6 & 70.3 & 87.3 & 82.0 & 87.6 & 31.1 & 79.0 & 61.9 & 81.6 & 80.4 & 84.5 & 83.3 & 58.4 & 86.1 & 55.9 & 78.2 & 65.4 \\
\hline
MoE-SPNet-CF & 74.1 &  90.3 & 40.1 & 81.9 & 62.4 & 70.9 & 90.3 & 87.5 & 88.4 & 33.7 & 81.1 & 56.3 & 82.5 & 83.0 & 87.0 & 83.6 & 57.2 & 85.2 & 50.0 & 83.0 & 66.9 \\
MoE-SPNet-EF & 74.2 &  89.2 & 38.8 & 79.1 & 64.1 & 72.8 & 90.9 & 87.0 & 88.6 & 35.2 & 81.9 & 61.2 & 83.7 & 80.3 & 84.5 & 83.5 & 59.5 & 83.9 & 55.6 & 78.3 & 67.5 \\
MoE-SPNet& 74.7 &  90.1 & 38.6 & 79.7 & 63.4 & 69.9 & 90.9 & 86.4 & 89.1 & 32.2 & 82.7 & 62.6 & 84.9 & 83.3 & 85.7 & 82.7 & 63.9 & 84.2 & 56.6 & 79.3 & 67.6 \\
\hline
\end{tabular}
\caption{\small{\textbf{Comparison of different MoEs for parsing.} Baseline: The DeepLab-ASPP parsing network without the gating part. MoE-SPNet-CF: The gating network using the input features to all the experts. MoE-SPNet-EF: The gating network takes the high-level features $\mathbb{F}$ within each expert as input. MoE-SPNet: The gating network takes the predictions $\mathcal{F}_i$ of each expert as input.}} 
\label{tab:v-MoEs}
\end{table}

Furthermore, Tab.~\ref{tab:v-MoEs} shows evaluation results of variants of the proposed MoE-SPNet on the test server, including MoE-SPNet-CF whose gating network share the same input with the that of the experts, MoE-SPNet-EF whose gating network takes the features within each experts as input. From the table, all the MoE based parsing networks yield at least an improvement of 1.0\% over the baseline network (Deeplab-ASPP). MoE-SPNet-EF outperforms MoE-SPNet-CF by 0.5\%, while MoE-SPNet obtains a further 0.6\% improvement compared with MoE-SPNet-EF, which demonstrate that direct understanding of a scene can help learn a more effective gating network. To exploit the MoE based parsing networks in deeper, we calculate the number of parameters of the gating networks belong to these MoE-SPNets here. Assuming the dimension of $\mathcal{S}$, $\mathbb{F}_{i}$, and $\mathcal{F}_{i}$ are $\mathcal{C}_{1}$,  $\mathcal{C}_{2}$,  and $\mathcal{C}_{3}$, respectively, and the additional convolutional layer for MoE-SPNet-EF and MoE-SPNet-CF contain $\mathcal{C}_{4}$ channels,
the numbers of the gating networks parameters in MoE-SPNet-CF, MoE-SPNet-CF, MoE-SPNet are $\mathcal{C}_{1} \times \mathcal{C}_{4} \times 3 \times 3 + \mathcal{C}_{4} \times \mathcal{N} \times 1 \times 1$, $\mathcal{N} \times \mathcal{C}_{2} \times \mathcal{C}_{4} \times 3 \times 3 + \mathcal{C}_{4} \times \mathcal{N} \times 1 \times 1$, and $\mathcal{N} \times \mathcal{C}_{3} \times \mathcal{N} \times 3 \times 3 $, respectively, where $\mathcal{N}$ is the number of experts, and $\mathcal{N} < \mathcal{C}_{3} \ll \mathcal{C}_{1}, \mathcal{C}_{2}, \mathcal{C}_{4} $. It can be seen that MoE-SPNet contains the fewest parameters but achieves the best performance by using a gating network learned from the predictions of each expert. 


\section{Conclusion}
In this paper, we have proposed MoE-SPNet and FCN-AHFA  to better exploit the diversities of contextual information in multi-level features and the spatial inhomogeneity of a scene in CNN-based models for scene parsing, by learning to assess the importance of features from different levels at each spatial location, instead of aggregating such features via concatenation or linear combination as commonly done in previous methods. The proposed MoE-SPNet achieves better performance by incorporating a mixture-of-experts layer to assess the importance of features from different layers. The AHFA scheme inspired by MoE-SPNet is applicable to a variety of scene parsing networks that use skip connections to fuse multi-level features from different stages. The value of the proposed methods have been demonstrated by the consistent and remarkable performance increase in a number of experiments on two challenging benchmarks (PASCAL VOC 2012 and SceneParse150).
In the future, we will continue investigating more effective and efficient methods to jointly make use of multi-level convolutional features in CNN-based models for scene parsing and other challenging computer vision problems.

\section*{References}

\bibliography{egbib}

\end{document}